\def\year{2020}\relax
\documentclass[letterpaper]{article} % DO NOT CHANGE THIS
\usepackage{aaai20}  % DO NOT CHANGE THIS
\usepackage{times}  % DO NOT CHANGE THIS
\usepackage{helvet} % DO NOT CHANGE THIS
\usepackage{courier}  % DO NOT CHANGE THIS
\usepackage[hyphens]{url}  % DO NOT CHANGE THIS
\usepackage{graphicx} % DO NOT CHANGE THIS
\usepackage{graphicx}  % DO NOT CHANGE THIS
\usepackage{multirow}
\usepackage{booktabs}
\title{Rethinking the Number of Channels for Convolutional Neural Networks}
\author{Hui Zhu\textsuperscript{\rm 1,2}, Zhulin An\textsuperscript{\rm 1}\thanks{Corresponding author of this work}, Chuanguang Yang\textsuperscript{\rm 1}, Xiaolong Hu\textsuperscript{\rm 1}, Kaiqiang Xu\textsuperscript{\rm 1}, Yongjun Xu\textsuperscript{\rm 1}\\ 
\textsuperscript{\rm 1}Institute of Computing Technology, Chinese Academy of Sciences\\
\textsuperscript{\rm 2}University of Chinese Academy of Sciences\\
\{zhuhui,anzhulin,yangchuanguang,huxiaolong,xukaiqiang,xyj\}@ict.ac.cn
}
\begin{document}

\maketitle

\begin{abstract}
 Latest algorithms for automatic neural architecture search perform remarkable but few of them can effectively design the number of channels for convolutional neural networks and consume less computational efforts. In this paper, we propose a method for efficient automatic architecture search which is special to the widths of networks instead of the connections of neural architecture. Our method, functionally incremental search based on function-preserving, will explore the number of channels rapidly while controlling the number of parameters of the target network. On CIFAR-10 and CIFAR-100 classification, our method using minimal computational resources (0.4$\sim$1.3 GPU-days) can discover more efficient rules of the widths of networks to improve the accuracy by about 0.5\% on CIFAR-10 and a$\sim$2.33\% on CIFAR-100 with fewer number of parameters. In particular, our method is suitable for exploring the number of channels of almost any convolutional neural network rapidly.
\end{abstract}

\section{Introduction}
In recent years, deep learning has achieved great success in the field of computer vision. As the crucial factor affecting the performance of convolutional neural networks, many excellent neural architectures have been designed, such as VGG \cite{vgg}, ResNet \cite{resnet}, DenseNet \cite{densenet}, SENet \cite{senet} and so on. Although these human-designed neural architectures constantly refresh the classification accuracy on specific datasets, they rely heavily on expert experience and it is extremely difficult to manually design suitable neural architectures when faced with brand new image tasks. With more and more attention paid to the research of automatic neural architecture search, many prominent neural architectures discovered by various kinds of neural architecture search algorithms \cite{AmoebaNet,enas,pnasreal,pnas,darts,proxylessnas}. Some of them have gradually surpassed the human-designed neural architectures in performance.

The choice of the number of channels has become an important research point since the initial human-designed convolutional neural networks were designed. Many classical convolutional neural networks, such as AlexNet \cite{alexnet}, VGG and ResNet usually sharply increase the feature map dimension (the number of channels) at downsampling locations in order to roughly increase the diversity of high-level attributes. Several subsequent researches have improved the number of channels for these thin and deep networks and achieve better results. For example, WRNs \cite{wrn} increase the widths of residual networks and PyramidNet \cite{pyramid} design the additive and multiplicative pyramid shapes to gradually increase the number of channels. With the continuous development of the filed of neural architecture search, many algorithms \cite{densenas,efficientnet} tend to seriously consider the number of channels for convolutional neural architectures. Although these methods have achieved outstanding results, they are all computational expensive and the search efficiency are not satisfactory.

In this paper, we propose a novel method based on function-preserving for efficient search for the width of neural networks and rethink the number of channels for convolutional neural networks. Function-preserving transformations derive from Net2Net \cite{net2net}, which target at transferring the information in the well-trained teacher neural networks into the student neural networks rapidly. This method has been well used in network morphism \cite{morphism} and several neural architecture search algorithms \cite{eas,ibm,eena}. Differently, we pay more attention on the number of channels instead of the connections of neural architectures. We propose the functionally incremental search which based on function-preserving to efficiently explore the number of channels of almost any network rapidly. We do confirmatory experiments on several classical convolutional neural networks and improve the performance of the original networks while controlling the number of parameters of the models. Based on the search results, we discuss the rules of the number of channels for convolutional neural networks in combination with a series of additional supplementary experiments.

The results of our experiments show that our search method achieves better classification accuracy (accuracy can be improved by about 0.5\% on CIFAR-10 and a$\sim$2.33\% on CIFAR-100) with fewer number of parameters by optimizing the number of channels for classical convolutional neural networks. The search process consumes approximately 0.4$\sim$1.3 GPU-days\footnotemark \footnotetext{All of our experiments were performed using a NVIDIA Titan Xp GPU.} computational resources which mainly depends on the complexity of the network.

Our contributions are summarized as follows:

\begin{itemize}
\item We propose an efficient neural architecture search method based on function-preserving, functionally incremental search, which may control the number of parameters and explore the number of channels for almost any convolutional neural network rapidly.
\item Based on the search results of our method and a series of additional supplementary experiments, we rethink the rules of the number of channels for convolutional neural networks.
\item We achieve better classification accuracy by modifying the widths of the original networks with remarkable architecture search efficiency which we attribute to the use of our method.
\end{itemize}

Part of the code implementation for search and several neural architectures we discovered on CIFAR-10 are available at \url{https://github.com/Search-Width/Search-the-Number-of-Channels}.

\section{Related Work}
In this section, we review the number of channels designed manually, function-preserving transformations and neural architecture search which are most related to this work.

\subsubsection{The Number of Channels Designed Manually.} For convolutional neural networks, appropriate depth and width choices may achieve higher accuracy and smaller model size. Even slight changes to the number of channels for the neural architecture will have a huge impact on the performance of the network. Based on ResNet \cite{resnet} blocks, Zagoruyko and Komodakis \cite{wrn} propose the wide residual networks (WRNs) which decrease depth and increase width of residual networks and show that these are far superior over their commonly used thin and very deep counterparts. In addition, for many classical convolutional neural networks such as VGG \cite{vgg} and ResNet, the number of channels is sharply increased at downsampling locations. It is considered to increase the diversity of high-level attributes and then roughly ensure effective performance of networks. Instead of this method, PyramidNet \cite{pyramid} gradually increase the feature map dimension at all units to involve as many locations as possible and the design has proven to be an effective means of improving generalization ability. From the existing researches as above, we can notice that convolutional neural networks can be improved by manually designing the number of channels but the method relies heavily on human experience.

\subsubsection{Function-Preserving Transformations.} Transferring a well-trained neural network to a new one with its network function completely preserved mainly contains two methods: Net2Net \cite{net2net} and network morphism \cite{morphism}. They can transfer the information stored in the teacher neural network into the student neural network rapidly based on function-preserving. Although they target at the same problem, there are several major differences between them. For example, Net2Net's operations only work for idempotent activation functions while network morphism may handle arbitrary non-linear activation functions. Based on function-preserving, we can quickly change the neural architecture while reducing redundant training. Thus, the search algorithm may explore the number of channels for convolutional neural networks with less search time and computational effort.

\subsubsection{Neural Architecture Search.} The latest algorithms for neural architecture search mainly include evolutionary algorithm (EA) \cite{hierarchical,AmoebaNet}, reinforcement learning (RL) \cite{nas,nasnet,enas}, Bayesian optimization (BO) \cite{bayesian} and gradient-based methods \cite{darts,naonet}. The algorithms of automatic neural architecture search are increasingly concerned with their own efficiency (such as search time) besides focusing on the effect of neural architectures discovered. Several methods \cite{rlnt,eas,eena} based on function-preserving perform prominent with less computational effort. However, existing methods basically pay less attention on the number of channels. Fang et al. \cite{densenas} propose a differentiable method called DenseNAS which can search for the width and the spatial resolution of each block simultaneously. In general, efficient search for the number of channels for convolutional neural networks is an intractable and poorly researched problem. In this work, we propose a novel method based on evolutionary algorithm and function-preserving transformations for efficient search specifically for the number of channels for convolutional neural networks and achieve remarkable results.

\section{Proposed Methods}
In this section, we review the approach to widen a convolutional layer based on function-preserving and illustrate our method of functionally incremental search in details.

\begin{figure*}[t]
  \centering
  \includegraphics[width=0.95\linewidth]{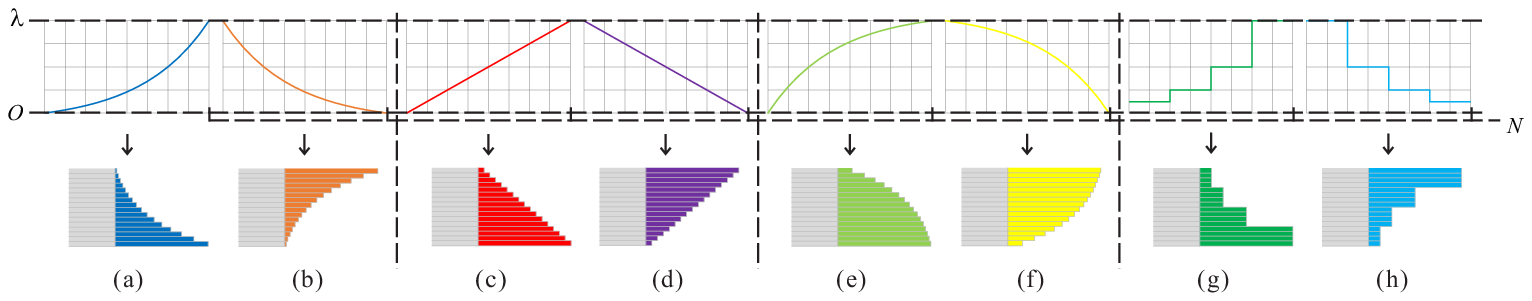}
  \caption{Visualization of the Functionally incremental search. The top half of (a) $\sim$ (h) are the 8 kinds of function to incrementally explore the number of channels and the bottom half are the schematic diagrams of the corresponding models after width changes. The range of each function is 0 $\sim$ $N$ on the horizontal axis and 0 $\sim$ $\lambda$ on the vertical axis.}
\end{figure*}

\subsection{Widening a Convolutional Layer.}
Assume that $x$ is the input to the network, function-preserving transformation is to choose a new set of parameters $\theta^{'}$ for a student network $G(x;\theta^{'})$ which transform from the teacher network $F(x;\theta)$ such that:
\begin{equation}
\forall x:F(x;\theta)=G(x;\theta^{'}).
\end{equation}
Assume that the $i$-th convolutional layer to be changed is represented by a $(k_{1},k_{2}, c, f)$  shaped matrix $W^{(i)}$. $W^{(i)}$ is extend by replicating the parameters along the last axis at random and the parameters in $W^{(i+1)}$ need to be divided along the third axis corresponding to the counts of the same filters in the $i$-th layer. $U$ is the new parameter matrix and $f^{'}$ is the number of filters in the layer $i$+1. Specifically, A noise $\delta$ is randomly added to every new parameter in $W^{(i+1)}$ to break symmetry. The operation for widening a layer can be expressed as follows:
\begin{equation}
g(j)=
\left\{  
	\begin{array}{lr}
	j  & j \leq f \\ 
	random \, sample \, from \lbrace 1,2,\cdots,f \rbrace  & j > f
	\end{array}
\right.,
\end{equation}
\begin{equation}
U_{k_{1},k_{2},c,j}^{(i)} = W_{k_{1},k_{2},c,g_{(j)}}^{(i)},
\end{equation}
\begin{equation}
U_{k_{1},k_{2},j,f^{'}}^{(i+1)} = \frac{ W_{k_{1},k_{2},g(j),f^{'}}^{(i+1)}}{\vert\{{x \vert g(x)=g(j)}\}\vert} \cdot {(1+ \delta)}, \, \delta \in \left[0, 0.05 \right].
\end{equation}

\subsection{Functionally Incremental Search.} The function for incremental search and the schematic diagrams of the corresponding models after width changes are visualized in Fig.1. Assume that $N$ represents the number of convolutional layers in the network. $\lambda$ represents the rate of increase, in other words, the number of channels will increase to $1+\lambda$ times at most. The range of the independent variable $x$ is $\left(0,N\right].$

\subsubsection{Function of Increasing Slop and Its Symmetric Function.} The function of increasing slop is shown in Fig.1(a) and it can be expressed as:
\begin{equation}
f_{a}\left( x \right) = \lambda\cdot\frac{\left(\lambda+1\right)^{x}-1}{\left(\lambda+1\right)^{N}-1} ,
\end{equation}
The function shown in Fig.1(b) is symmetric with the function in eq.(5) about $x=N/2$. It can be expressed as:
\begin{equation}
f_{b}\left( x \right) = \lambda\cdot\frac{\left(\lambda+1\right)^{N-x}-1}{\left(\lambda+1\right)^{N}-1}.
\end{equation}
Thus, after the incremental changes according to these two functions, the number of channels for the convolutional layers in the network are opposite. The following symmetric functions have the same effects.

\subsubsection{Function of Constant Slop and Its Symmetric Function.} The function of constant slop is shown in Fig.1(c) and it can be expressed as:
\begin{equation}
f_{c}\left( x \right) = \frac{\lambda}{N}x,
\end{equation}
Similar to the above, the function shown in Fig.1(d) is symmetric with the function in eq.(7) about $x=N/2$. It can be expressed as:
\begin{equation}
f_{d}\left( x \right) = \lambda - \frac{\lambda}{N}x.
\end{equation}

\subsubsection{Function of Decreasing Slop and Its Symmetric Function.} The function of decreasing slop is shown in Fig.1(e). It is symmetric with the function in eq.(6) about $f(x)=\lambda/2$ and can be expressed as:
\begin{equation}
f_{e}\left( x \right) = \lambda\cdot\frac{\left(\lambda+1\right)^{N}-\left(\lambda+1\right)^{N-x}}{\left(\lambda+1\right)^{N}-1},
\end{equation}
Similar to the above, the function shown in Fig.1(f) is symmetric with the function in eq.(9) about $x=N/2$. It can be expressed as:
\begin{equation}
f_{f}\left( x \right) = \lambda\cdot\frac{\left(\lambda+1\right)^{N}-\left(\lambda+1\right)^{x}}{\left(\lambda+1\right)^{N}-1}.
\end{equation}

\subsubsection{Piecewise Function of the Step Shape.} Assume that $K_{i}$ represents the number of convolution layers before the $i$-th downsampling operation. Set $n = max\{i\}+1$, the function of the Step Shape is shown in Fig.1(g) and it can be expressed as:
\begin{equation}
f_{g}\left( x \right) =
\left\{  
	\begin{array}{lll}
	\frac{1}{2^{n-1}}\cdot\lambda &\quad & \,0 < x \leq K_{1} \smallskip\\ 
	\frac{1}{2^{n-2}}\cdot\lambda &\quad & K_{1} < x \leq K_{2} \smallskip\\
	\, \cdots &\quad & \, \cdots \smallskip\\
	\,\lambda &\quad & K_{n-1} < x \leq N
	\end{array}
\right.,
\end{equation}
The function shown in Fig.1(h) is the opposite of the function in eq.(11) and it can be expressed as:
\begin{equation}
f_{h}\left( x \right) =
\left\{  
	\begin{array}{lll}
	\,\lambda  &\quad & \,0 < x \leq K_{1} \smallskip\\ 
		\, \cdots &\quad & \, \cdots \smallskip\\
	\frac{1}{2^{n-2}}\cdot\lambda &\quad & K_{1} < x \leq K_{2} \smallskip\\
	\frac{1}{2^{n-1}}\cdot\lambda &\quad & K_{n-1} < x \leq N
	\end{array}
\right..
\end{equation}
Notice that these two functions are not necessarily symmetric about $x=N/2$ as the number of convolution layers may be different for every two intervals divided by the downsampling operation.

\begin{figure}[t]
  \centering
  \includegraphics[width=0.99\linewidth]{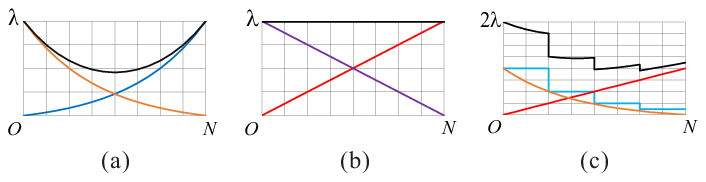}
  \caption{Several complex rules of the width for the network may be discovered by calculating the increments by the initial fixed number of channels. The black curves represent the number of channels after multiple changes.}
\end{figure}

\subsubsection{Calculate the Increment of the Widths.} The methods used to calculate increments mainly include: Calculating by the initial fixed number of channels and calculating by the number of channels that may change during the search process. For the former, Each calculation is based on the initial network in which the number of channels will be unchanged. Several examples of this are visualized in Fig.2. As shown in Fig.2(a), the width of the network may decrease and then increase after multiply changes. In Fig.2(b) and Fig.2(c), the number of channels per layer may become the same or become unpredictably  complex. In this case, several functions (e.g. $f\left(x\right)=\lambda$ in Fig.2(b)) become useless because their effects can be generated by the addition of other functions. For the latter, Each calculation is based on the new network in which the number of channels is changed per round. Assume that $\Theta$ is the initial set of the convolutional layers in the network and the width of the $i$-th layer is represented as $\theta_{i}$. $\Theta^{'}$ is the set after $n$ changes and $\xi$ represents a random selected function. It can be expressed as: 
\begin{equation}
\Theta=\left\{\theta_{i}\right\}, \quad \Theta^{'}=\left\{\theta_{i}\cdot \prod_{j=1}^{n} \left(1+f_{\xi}\left( i \right)\right) \right\}.  
\end{equation}
In this case, we may add the function $f\left(x\right)=\lambda/2$ to ensure that the network with the same number of channels in all convolutional layers can be produced. Various complex rules of the number of channels may be discovered gradually through the above two methods. For the sake of search efficiency, we prefer the latter one in our experiments.

\subsection{Search Based on Evolutionary Algorithm.} Our search method is based on the evolutionary algorithm. The initial network is randomly mutated several times to form the initial population of size $P_{1}$. Then we use tournament selection \cite{tournament} to select an individual for mutation: a fraction $k$ of individuals is selected from the population randomly and the individual with highest fitness is final selected from this set. After several rounds, the population size will reach $P_{2}$. From this point on, the individual with the lowest fitness in the population will be discard while a new individual is generated through mutation. Thus, the population size remains unchanged ($P_{2}$) until the end of the evolution.

\section{Experiments}
In this section, we introduce the implementation of experiments and report the performances of functionally incremental search. The experiments are mainly implemented in accordance with the methods mentioned in Sect.3 and we make a detailed introduction here. In addition, we compare the networks discovered with the original classical ones to prove the feasibility and efficiency of our method.

\subsection{Initial Model.} The method for constructing the initial model is described in Fig.3. The widths of the original network are usually varied as shown on the left and we choose the same number of channels for every convoluaional layer as shown on the right in the schematic diagram. The main purpose of this is to create a large and free enough search space and then we can use our method to discover the widths of networks that may be more suitable for the given datasets. In other words, We will select a small number of channels (usually half of the minimum number of channels) from the classical convolutional neural network for search. For example, the number of channels for ResNet-18 includes 64$\times$4, 128$\times$4, 256$\times$4 and 512$\times$4. We will construct a small network with the same depth (16 convolutional layers) but the number of channels are all initialized to 32. The weights are initialized as He normal distribution \cite{resnet} and the L2 regularization of 0.0001 is applied to the weights.

\begin{figure}[t]
  \centering
  \includegraphics[width=0.85\linewidth]{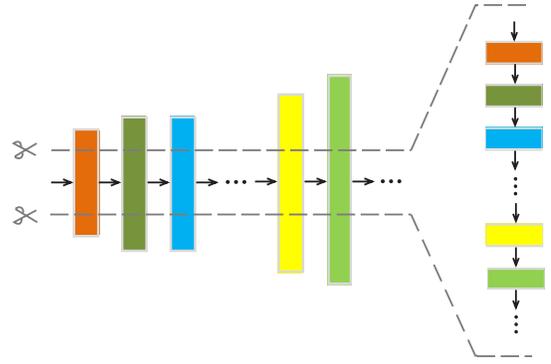}
  \caption{Select a small number of channels from the classical convolutional neural network to form the initial model for search. Different colored rectangles represent different convolutional layers.}
\end{figure}

\subsection{Dataset.} For CIFAR-10 and CIFAR-100, We randomly sample 10,000 images by stratified sampling from the original training set to form a validation set for evaluate the fitness of the individuals while using the remaining 40,000 images for training the individuals during the evolution. We normalize the images using channel means and standard deviations for preprocessing and apply a standard data augmentation scheme (zero-padding with 4 pixels on each side to obtain a $40\times40$ pixels image, then randomly cropping it to size $32\times32$ and randomly flipping the image horizontally). 

\begin{figure*}[t]
  \centering
  \includegraphics[width=0.9\linewidth]{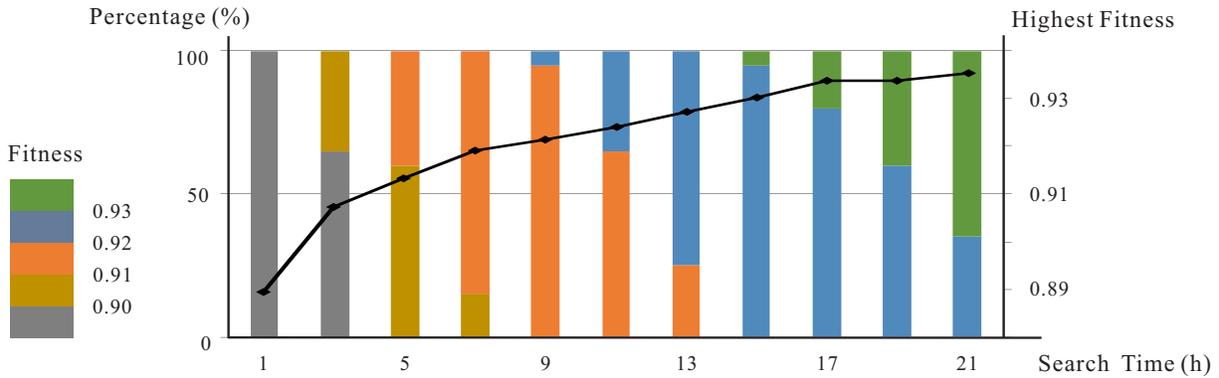}
  \caption{The schematic diagram visualized one search process of ResNet-18 on CIFAR-10. The horizontal axis represents the search time. The vertical axis on the left represents the percentage of individuals in each range of fitness and that on the right represents the highest fitness of the population. The bar chart represents the percentage and the line chart represents the change of the highest fitness.}
\end{figure*}

\subsection{Search for the Number of Channels.} 
We search for the number of channels with the evolution algorithm mentioned in Sect.3.3. In our experiments, A mutation represents a functionally incremental change. As mentioned in Sect.3.2, each calculation is based on the new network in which the number of channels is changed per round. So we add the function $f\left(x\right)=\lambda/2$ besides the other 8 functions and the probability of choosing each mutation is the same from the beginning to the end. Here, we describe the general process of search and the implementation details of different networks respectively.

\subsubsection{The General Process of Search.} In the process of search, the rate of increase $\lambda$ is fixed to 0.2. The initial population consists of 12 individuals, each formed by a single mutation from the initial model. The size of k in tournament selection is fixed to 3 and we start to discard individuals when the population size grows to 20. In other words, we discard the individual with the lowest fitness while we select an individual to mutate and put the new individual back into the population after the population size reaches 20. All the networks are trained with a batch size of 128 using SGDR \cite{SGDR} with Nesterov's momentum set to 0.9, initial learning rate $l_{max}=0.05$, $T_{0}=1$ and $T_{mult}=2$. But the initial models will be trained for different epochs (usually 31 epochs with SGDR, a few may start with additional 10 epochs with SGD) because it takes more epochs to train a large-scale network to convergence. The individuals generated by mutation are trained for 15 epochs and then we calculate the fitness (accuracy) on the validation dataset. The process is repeated until the number of parameters of the individual with the highest fitness is similar to the comparison network. One search process of ResNet-18 on CIFAR-10 is visualized in Fig.4. The horizontal axis represents the search time. The vertical axis on the left represents the percentage of individuals in each range of fitness as shown in illustration. The vertical axis on the right represents the highest fitness of the population. The bar chart represents the percentage and the line chart represents the change of the highest fitness. Although the process of evolution will slow down gradually, we can notice that the fitness of the population increases steadily and rapidly via our method. Finally, the individual with the highest fitness will be selected for more training and be used to compare with the original network. Specifically, we search ResNet-18 for several times while other classical convolutional neural networks just once.

\begin{table*}[t]
  \caption{Comparison against the results of classical networks on CIFAR-10 and CIFAR-100. To enable a fair comparison, we retrain the original networks by the same training method with the best networks discovered by our search. After modifying the number of channels for the classical convolutional neural networks, the number of parameters and the classification error are both reduced. The numbers in brackets denote the performance improvement over the original baselines.}\smallskip \smallskip
  \centering
  \resizebox{1.9\columnwidth}{!}{
  \begin{tabular}{l||ccc||cccc}
    \toprule
    \midrule
    \multirow{3}{*}{\textbf{Method}} & \multicolumn{3}{c||}{\textbf{Original networks}} & \multicolumn{4}{c}{\textbf{Modify the Number of Channels}} \\
    \cmidrule{2-8}
    & \textbf{Params} & \textbf{C10 Test Error} & \textbf{C100 Test Error} & \textbf{Params} & \textbf{C10 Test Error} & \textbf{C100 Test Error} & \textbf{Search Time} \\
     & (Mil.) & (\%) & (\%) & (Mil.) & (\%) & (\%) & (GPU-days) \\
    \midrule
    \multirow{3}{*}{ResNet-18  \cite{resnet}} & \multirow{3}{*}{11.54}  & \multirow{3}{*}{3.86} & \multirow{3}{*}{22.76} & 6.98 & $3.69_{(0.17)}$ & $20.60_{(2.16)}$ & 0.48   \\
    & & & & 9.94 & $3.40_{(0.46)}$ & $20.74_{(2.02)}$ & 0.92 \\
    & & & & 10.57 & $3.43_{(0.43)}$ & $20.46_{(2.30)}$ & 0.66 \\
    \midrule
    ResNet-34 \cite{resnet}  & 22.22  & 4.71 & 24.81 & 13.00 & $4.39_{(0.32)}$ & $24.44_{(0.37)}$ & 1.29   \\
    \midrule
    VGG-16 \cite{vgg}  & 15.00  & 5.95 & 28.38 & 7.24 & $5.43_{(0.52)}$ & $27.85_{(0.53)}$ & 0.41   \\
    \midrule
    SE-ResNet-50 \cite{senet} & 26.12 & 3.81 & 22.30 & 11.96 & $3.38_{(0.43)}$ & $19.97_{(2.33)}$ & 0.79  \\
     \midrule
    \bottomrule
  \end{tabular}
  }
\end{table*}

\subsubsection{Implementation Details of ResNet-18.} For ResNet-18, we train the initial models for 31 epochs with SGDR. As the feature map dimension of the network is increased at every unit, we use projection shortcuts conducted by $1\times1$ convolutions. The two convolution layers of a residual block are changed separately, that is, there is no identity. We conduct an additional experiment to show that this operation will not lead to lower accuracy of the network. We train the network whose feature map dimension is increased at every unit and the original ResNet for 511 epochs respectively. The accuracy of the two networks are almost the same (the former may be about 0.05\% higher).

\subsubsection{Implementation Details of ResNet-34.} For ResNet-34, we train the initial models for 10 epochs with SGD and then 31 epochs with SGDR. As the feature map dimension of the network is increased at every unit, we also use projection shortcuts conducted by $1\times1$ convolutions. Although it will not affect the comparison of our method, this operation may lead to slightly lower accuracy than the original ResNet-34. On the one hand, Cutout and SGDR may have some effect. On the other hand, the crucial cause of this problem has been discussed by He et al. \cite{identity-mapping}. A projection shortcut can hamper information propagation and lead to optimization problems, especially for very deep networks.

\subsubsection{Implementation Details of VGG-16.} For VGG-16, there are several differences with the original VGG-16 and other networks. We change the the number of fully-connected layers to 2, which contain a hidden layer with 512 units and a softmax layer. Dropout layers are added to the convolutional layers ($drop\_rate=0.3$ or $0.4$) and fully-connected layers ($drop\_rate=0.5$). The weights are initialized as Xavier uniform distribution and the L2 regularization of 0.0005 is applied to the weights. We train the initial model for 30 epochs with SGD and then 31 epochs with SGDR.

\subsubsection{Implementation Details of SE-ResNet-50.} For SE-ResNet-50, we train the initial model for 10 epochs with SGD and then 31 epochs with SGDR. Not exactly the same as ResNet-34, we use projection shortcuts conducted by $1\times1$ convolutions but we take the  layers in an interval separated by the downsampling operation as a whole. In addition, the 3 convolutional layers of the bottleneck block are identical and the ratio of the number of channels is fixed to 1:1:4.

\subsection{Training Methods and Results.} 
The training methods we mentioned here refer to the methods used for finally training the networks to convergence for comparison of their performances. We used exactly the same training methods to train the original classical networks and the best network discovered by our functionally incremental search. Based on the training results, the comparative analysis proves the effectiveness of our method. 

\subsubsection{Training Methods.} Networks are trained on the full training dataset until convergence using Cutout \cite{cutout}. All the networks are trained with a batch size of 128 using SGDR \cite{SGDR} with Nesterov's momentum for 511 epochs (several networks may be trained for several epochs with SGD first, such as ResNet-34 and VGG-16). The hyper-parameters of the methods are as follows: the cutout size is $16\times16$ for Cutout, $momentum=0.9$, $l_{max}=0.1$,$T_{0}=1$ and $T_{mult}=2$ for SGDR. Finally, the error on the test dataset will be reported.

\subsubsection{Comparison with Classical Networks.} The comparison against the results of original classical networks on CIFAR-10 and CIFAR-100 is presented in Table 1. We show the comparison of the number of parameters and the test error on the datasets. The numbers in brackets denote the performance improvement over the original baselines. In addition, we add the search time for our method to show the efficiency of the search. Specifically, the number of parameters of the networks used to compare the test error on CIFAR-100 are slightly more than on CIFAR-10 because of the last fully-connected layer (10-way and 100-way). Since the number of parameters are almost the same, we only show the one for CIFAR-10 in the table. Our method is suitable for exploring the number of channels of almost any network rapidly and we select several classical networks for experiments. We can notice that our method using minimal computational resources (0.4$\sim$1.3 GPU-days) can discover more efficient widths of networks (the accuracy can be improved by about 0.5\% on CIFAR-10 and 0.37\%$\sim$2.33\% on CIFAR-100 with fewer number of parameters).

\begin{table}[t]
  \caption{Comparison of results for different rules of the number of channels for ResNet-18 on CIFAR-10. The first three rows are based on the original ResNet and the last two rows are based on the number of channels we discovered by functionally incremental search for ResNet.}\smallskip \smallskip
  \centering
  \resizebox{1\columnwidth}{!}{
  \begin{tabular}{l|c|c}
    \toprule
    \midrule  
    \multirow{2}{*}{\textbf{Method}} & \textbf{Params} &  \textbf{C10 Test Error} \\
    & (Mil.) & (\%)\\
	\midrule
    Increasing number of channels (Original ResNet-18)   & 11.18 &  3.88   \\
    \midrule
    Constant number of channels (Original ResNet-18)   & 9.23 &   3.60  \\
    \midrule
     Decreasing number of channels (Original ResNet-18)   & 11.47 &  3.83   \\
    \midrule
    Increasing number of channels (Modified ResNet-18)   & 9.94 &  \textbf{3.40}   \\
    \midrule
    Decreasing number of channel (Modified ResNet-18)    & 9.96 &   3.54  \\
    \midrule   
    \bottomrule
  \end{tabular}
  }
\end{table}

\begin{table}[t]
  \caption{Comparison of results for different rules of the number of channels for PyramidNet-110 on CIFAR-10. The first row is based on the original PyramidNet and the last row is designed according to the rule of the number of channels we discovered by functionally incremental search.}\smallskip \smallskip
  \centering
  \resizebox{1\columnwidth}{!}{
  \begin{tabular}{l|c|c}
    \toprule
    \midrule  
    \multirow{2}{*}{\textbf{Method}} & \textbf{Params} &  \textbf{C10 Test Error} \\
    & (Mil.) & (\%)\\
	\midrule
    Increasing number of channels (Original PyramidNet-110)   & 3.88 &  3.77   \\
    \midrule
    Constant number of channels (Modified PyramidNet-110)   & 3.79 &   3.65  \\
    \midrule
    fluctuating number of channels (Modified PyramidNet-110)   & 3.86 &  \textbf{3.54}   \\
    \midrule
    \bottomrule
  \end{tabular}
  }
\end{table}

\subsection{Rethinking the number of channels.} 
Based on the rules of the widths searched by our method for different convolutional neural networks, we can notice that the number of channels discovered are quite different from those designed manually by the predecessors. To further analyse the rules of the number of channels, we conduct a series of additional supplementary experiments for ResNet-18 and PyramidNet-110 on CIFAR-10.

\subsubsection{Additional Supplementary experiments.} We change the original (increasing) number of channels for ResNet-18 to constant (256$\times$16) and decreasing (512$\times$4,256$\times$4,128$\times$4,64$\times$4) respectively. In addition, we change the number of channels we discovered by functionally incremental search for ResNet-18 to the opposite order (the second row in Table 1 with the results searched for 0.92 GPU$-$days). Then we use the same methods mentioned above to train the networks to convergence. The comparison of results for different rules of the number of channels for ResNet-18 on CIFAR-10 is presented in Table 2.  For PyramidNet-110, we also conduct a series of similar experiments. We select the original additive PyramidNet (the performance is slightly better than multiplicative PyramidNet) whose widening factor $\alpha=84$ and the feature map dimension of the first convolutional layer is 16. In order to avoid the problems brought by projection shotcuts, we consider the option used in the original PyramidNet: zero-padded identity mapping shortcuts. In other words, we only use projection shotcuts when decreasing the feature map dimensions. As comparisons, we change the original number of channels for PyramidNet to constant (62$\times$110) and that designed according to the rule of the number of channels we discovered by our method for ResNet. The training methods are the same and the comparison of results for different rules of the number of channels for PyramidNet-110 on CIFAR-10 is presented in Table 3. 

\subsubsection{Discussions on the number of channels.} 
Combined with our previous experiment results, we can notice that the performance of increasing the feature map dimension sharply at downsampling locations is not very excellent. It is even possible that the most suitable rule of the number of channels for the convolutional neural networks on a particular dataset is not necessarily continuously increasing. Surprisingly, the decreasing number of channels may achieve better performance if only the number of parameters is taken into account without considering the amount of computation (FLOPs). Fluctuating widths of the networks can perform better than that sharply increase the widths at downsampling locations and that gradually increase the feature map dimension at all units to involve as many locations as possible. It seems like a good solution to use our search method to explore a complex and fluctuating rule of the number of channels suitable for a given dataset.

\section{Conclusions}
We proposed an efficient method based on function-preserving to search the number of channels for convolutional neural networks. The classical convolutional networks with the widths explored by the functionally incremental search perform better than the original ones. Obviously, The networks modified by our method contain fewer parameters and achieve higher accuracy on the CIFAR-10 and CIFAR-100. We notice that the amount of computation (FLOPs) is also an important part of the performance of networks. Therefore, we are trying to add the number of parameters and amount of computation into the fitness evaluation function besides the accuracy.

\section*{Appendix}
Here we publish the number of channels used in our tables for the original convolutional neural networks and the ones searched by our methods. In particular, We change the number of channels which is an odd number to an adjacent even number (increase) and the the number of channels for PyramidNet is designed according to other rules rather than automatic search.

\begin{table}[t]
  \caption{The number of channels used to compare the classical convolutional neural networks and the networks modified by our method.}\smallskip \smallskip
  \centering
  \resizebox{1\columnwidth}{!}{
  \begin{tabular}{l|c|c}
    \toprule
    \midrule  
    \textbf{Method} & \textbf{Original networks} & \textbf{Modify the Number of Channels} \\
	\midrule
    \multirow{2}{*}{ResNet-18}   & \multirow{2}{*}{64$\times$4,128$\times$4,256$\times$4,512$\times$4}  & 198,200,210,216,192,194,202,208,202,200,216,218,244,254,272,284    \\
    \cmidrule{3-3}
    & & 200,206,226,238,212,228,242,290,258,256,280,280,286,314,320,324 \\
    \cmidrule{3-3}
    & & 248,272,304,336,256,272,292,298,252,264,264,266,244,248,236,230 \\
    \midrule
    \multirow{2}{*}{ResNet-34}   & \multirow{2}{*}{64$\times$6,128$\times$8,256$\times$12,512$\times$6}  & \,474,420,364,330,304,280,222,208,202,192,186,178,170,166,172,166,  \\
    & & 166,160,152,144,140,140,136,136,134,132,136,126,128,128,126,116 \\
    \midrule
    VGG-16  & 64$\times$2,128$\times$2,256$\times$3,512$\times$6  &  178.176,214,220,228,230,234,236,302,304,308,316,320 \\
    \midrule
    \multirow{2}{*}{SE-ResNet-50} & \{64,64,256\}$\times$3,\{128,128,512\}$\times$4,  & \,\{146,146,584\}$\times$3,\{154,154,616\}$\times$4,  \\
    & \{256,256,1024\}$\times$6,\{512,512,2048\}$\times$3 & \{200,200,800\}$\times$6,\{222,222,888\}$\times$3 \\
    \midrule   
    \midrule 
   PyramidNet-110  & \{16,17,19,20,22,23,25,$\cdots$,100\}$\times$2  &  60$\times$2,\{62,70,78,58,62,66,68,56,60,60,60,56,56\}$\times$8,\{54\}$\times$4 \\
    \midrule  
    \bottomrule
  \end{tabular}
  }
\end{table}

\bibliographystyle{aaai}
\bibliography{ref}

\end{document}